\documentclass[11pt]{article}

\usepackage[final]{acl}

\usepackage{times}
\usepackage{latexsym}

\usepackage[T1]{fontenc}
\usepackage[utf8]{inputenc}
\usepackage{microtype}
\usepackage{inconsolata}
\usepackage{graphicx}
\usepackage{xspace}
\usepackage{tabularray}
\UseTblrLibrary{booktabs}
\usepackage{makecell}
\usepackage{amsmath}
\usepackage{amsfonts}
\usepackage{listings}
\newcommand{\projectname}{\textsc{Customer-R1}\xspace}
\lstdefinestyle{promptstyle}{
    basicstyle=\small\ttfamily,
    breaklines=true,
    frame=single,
    captionpos=b,
    keepspaces=true,
    numbers=none,
    showspaces=false,
    showstringspaces=false,
    showtabs=false,
    tabsize=2,
    backgroundcolor=\color{gray!10},
    columns=flexible,
    breakatwhitespace=true,
    postbreak=\mbox{\textcolor{red}{$\hookrightarrow$}\space},
}

\title{Customer-R1: personalized simulation of Human Behaviors via RL-based LLM Agent in Online Shopping}

\title{Customer-R1: Personalized Simulation of Human Behaviors via RL-based LLM Agent in Online Shopping}

\author{
 \textbf{Ziyi Wang\textsuperscript{1}},
 \textbf{Yuxuan Lu\textsuperscript{1}},
 \textbf{Yimeng Zhang\textsuperscript{2}},
 \textbf{Jing Huang\textsuperscript{3}},
 \textbf{Dakuo Wang\textsuperscript{1}},
\\
 \textsuperscript{1}Northeastern University,
 \textsuperscript{2}Michigan State University,
 \textsuperscript{3}Amazon,
\\
 \small{
   \textbf{Correspondence:} \href{mailto:wang.ziyi19@northeastern.edu}{wang.ziyi19@northeastern.edu}, \href{mailto:d.wang@northeastern.edu}{d.wang@northeastern.edu}
 }
}

\begin{document}
\maketitle
\begin{abstract}
Simulating step-wise human behavior with Large Language Models (LLMs) has become an emerging research direction, enabling applications in various practical domains.
While prior methods, including prompting, supervised fine-tuning (SFT), and reinforcement learning (RL), have shown promise in modeling step-wise behavior,
they primarily learn a population-level policy without conditioning on a user’s persona, yielding generic rather than personalized simulations.
In this work, we pose a critical question: how can LLM agents better simulate personalized user behavior?
We introduce \projectname, an RL-based method for personalized, step-wise user behavior simulation in online shopping environments. 
Our policy is conditioned on an explicit persona, and we optimize next-step rationale and action generation via action correctness reward signals.
Experiments on the OPeRA dataset demonstrate that \projectname not only significantly outperforms prompting and SFT-based baselines in next-action prediction tasks, but also better matches users' action distribution, indicating higher fidelity in personalized behavior simulation.
\end{abstract}

\section{Introduction}
\begin{figure}[t]
    \centering
    \includegraphics[width=\linewidth]{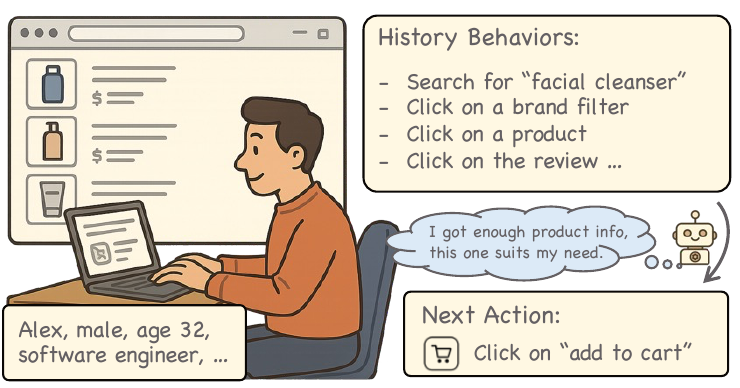}
    \caption{User Behavior Simulation in Online Shopping. The model observes a sequence of historical user actions and learns to reason over this behavioral context to predict the user’s next action. }
    \label{fig:page1_pic}
\end{figure}
 
Human behavior simulation~\citep{park2023generative, park2024generative, lu2025prompting} aims to model how humans take actions. Recent advances in the reasoning capabilities~\citep{yao2023react,shinn2023reflexion} of Large Language Model Agents (LLM Agents) have enabled both believable~\citep{park2023generative} and accurate~\citep{park2024generative} simulations of human behavior, drawing increasing attention to this topic.
These advancements have opened up new application opportunities in various practical domains, including computational social science~\citep{park2023generative}, psychology~\citep{10.5555/3618408.3618425}, e-commerce~\citep{lu2025prompting}, and usability testing~\citep{lu2025uxagentllmagentbasedusability}.

Recent efforts in human behavior simulation have shifted from simulating coarse-grained or unverified human behaviors towards accurately modeling step-wise actions~\cite{lu2025prompting,wang2025operadatasetobservationpersona,zhang2025shopr1rewardingllmssimulate}. 
However, existing methods typically learn an average-user policy: they predict the most common next action in seen contexts, but fail to account for individual differences in goals, preferences, or browsing styles. This lack of personalization limits their usefulness since different users may take very different actions in the same context~\cite{helmi2023characterizing}.
For instance, \citet{lu2025prompting} introduced step-wise user behavior simulation via supervised fine-tuning (SFT) on private data. \citet{zhang2025shopr1rewardingllmssimulate} proposed Shop-R1, a Reinforcement Learning (RL)-based approach to improve action generation accuracy. While promising, neither method is conditioned on individual user traits or preferences. 
Although \citet{wang2025operadatasetobservationpersona} benchmarked the value of persona information in the OPeRA dataset, they only evaluated the prompting method with off-the-shelf LLMs, which yields marginal gains in aligning actions to a specific user. These gaps motivate our question: 
how can LLM agents better simulate personalized user behavior?

To study this question systematically, we introduce \projectname, a reinforcement learning-based method for step-wise and personalized user behavior simulation in online shopping scenarios. An overview of the task setup is illustrated in Figure~\ref{fig:page1_pic}. The model takes in a sequence of historical user actions taken by user `Alex', and learns to reason over the behavioral context to predict Alex's next action accordingly. Our method leverages explicit user persona information to guide the model toward individualized behavioral patterns and introduces a tailored reward design to encourage accurate and semantically coherent action generation. We conduct extensive experiments on the OPeRA dataset~\citep{wang2025operadatasetobservationpersona}, which includes rich user interaction logs and annotated persona profiles. Results show that \projectname significantly outperforms prompting-based and SFT-based baselines in next-action prediction task and exhibits more aligned action distribution with persona information. Ablation studies further confirm the importance of persona conditioning: using correct persona information improves performance, while shuffled personas introduce noise and degrade accuracy.
The contributions of this work are as follows:

1) We introduce \projectname, a reinforcement learning-based method for personalized, step-wise user behavior simulation in online shopping, incorporating explicit persona information and custom reward design.

2) We provide a comprehensive evaluation on the OPeRA dataset, demonstrating substantial improvements over existing methods.

3) We conduct detailed ablations and analysis on persona, rationale, model size, and context length, together with error studies. These results offer practical guidance for building personalized behavior simulators in online shopping.

\section{Related Works}
\subsection{Human Behavior Simulation with Large Language Models}

Understanding and simulating human behavior has long been a central goal in psychology, human-computer interaction, and computational social science~\citep{mcconnell1974understanding,10.1145/3708359.3712149}. The emergence of large language model (LLM) agents with human-like reasoning, planning, and tool-use abilities~\citep{anthropic2025claude37,yao2023react,shinn2023reflexion} has opened new opportunities for modeling complex behaviors across diverse environments~\cite{wang2024userbehaviorsimulationlarge,lu2025uxagentllmagentbasedusability,wang2025agentabautomatedscalableweb}.
For example, in computational social science, generative agents have been used to simulate daily routines and social interactions in virtual communities~\citep{park2023generative}. 
However, many of these efforts primarily focus on generating ``believable'' user behavior, without quantitative evaluation against real human data. A few studies, such as~\citet{lu2025prompting}, have explored step-wise behavior simulations in online shopping and evaluated next-action prediction using real-world user traces. Yet these approaches often rely on private datasets and supervised fine-tuning (SFT), limiting reproducibility and further generalization. More recently, \citet{zhang2025shopr1rewardingllmssimulate} applied reinforcement learning (Shop-R1) to further improve action generation. However, these works focus on simulating an ``average'' user instead of an unique individual.

In terms of personalization, recent studies have begun incorporating user personas into behavior simulation~\citep{sun2024persona,park2024generative,wang2025operadatasetobservationpersona}. For instance,~\citet{park2024generative} showed that agents equipped with interview-based persona profiles exhibit improved performance in survey-taking tasks.~\citet{wang2025operadatasetobservationpersona} also introduced persona into simulations, but the experiments are conducted solely on off-the-shelf LLMs without task adaptation, showing limited performance gain. As a result, verifiable and personalized user simulation at the action level remains underexplored. To fill the gap, this study investigate how user persona information can enhance personalized behavior simulation.

\begin{figure*}[t]
    \centering
    \includegraphics[width=1.0\linewidth]{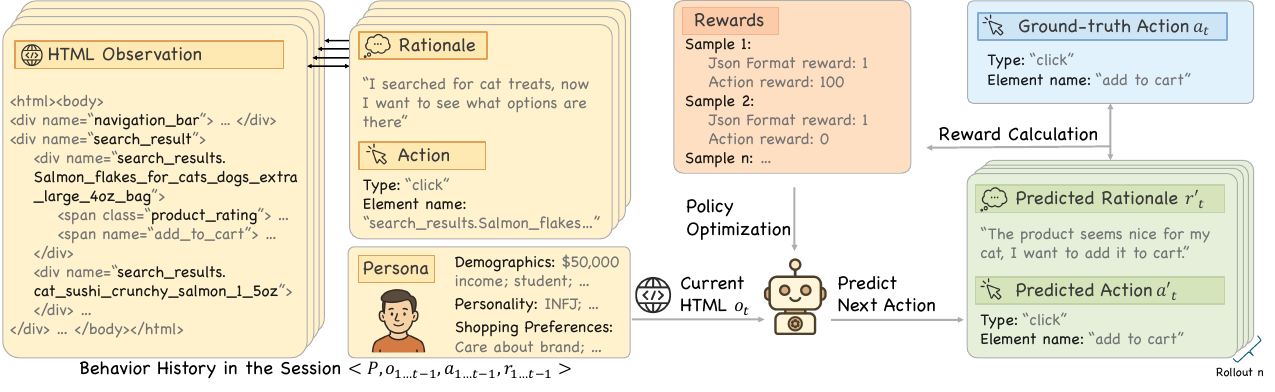}
    \caption{\textbf{\projectname Framework for Simulating User Behavior in Online Shopping.} 
        The model observes user history behaviors in a session composed of HTML observations $o_1, \dots,o_{t-1}$, actions $a_1, \dots,a_{t-1}$, rationales $r_1, \dots,r_{t-1}$, along with real user persona $P$ (demographics, personality, and shopping preferences). At time step $t$, given the current HTML observation $o_t$, the model predicts the rationale $r'_t$ for conducting an action and the corresponding next action $a'_t$. During training, the model samples $n$ rollouts per step. For each sampled prediction, a reward is calculated by comparing the predicted action $a'_t$ with the ground-truth action $a_t$ based on action correctness and format validity. These rewards are aggregated and used for policy optimization.
    }
    \label{fig:framework}
\end{figure*}

\subsection{Reinforcement Learning for LLM Post-Training}
Reinforcement learning (RL) has emerged as a powerful approach for training large language models (LLMs). Early methods, such as PPO~\cite{schulman2017proximalpolicyoptimizationalgorithms}, RLHF~\cite{ouyang2022traininglanguagemodelsfollow}, and DPO~\cite{rafailov2024directpreferenceoptimizationlanguage} focus on aligning model outputs with human or proxy preferences. More rencently, methods with verifiable reward signals, such as GRPO~\citep{deepseekai2025deepseekr1incentivizingreasoningcapability}, DAPO~\cite{yu2025dapoopensourcellmreinforcement}, and GSPO~\cite{zheng2025groupsequencepolicyoptimization}, have further improved stability and scalability. 
Furthermore, \citet{chen2025sft} systematically explores the contrast between supervised fine-tuning (SFT) and RL-based training paradigms, showing that GRPO-based methods can elicit stronger reasoning abilities compared to traditional SFT approaches.
Despite these advancements, much of current RL application targets tasks with clear correctness criteria, such as mathematical problem solving~\citep{shao2024deepseekmath,zhang2025r1}. In these settings, reward design is straightforward because binary or graded notions of correctness are available.
Extending RL to open-ended or user-centric tasks still poses challenges, particularly in defining meaningful and stable reward signals. In response, recent efforts have explored RL for more complex language interaction settings. For instance, RL has been applied to improve retrieval-augmented question answering systems~\citep{jin2025search} and recommender system outputs~\citep{lin2025rec}, where reward signals must account for relevance, diversity, or user engagement.~\citet{wei2025webagentr1trainingwebagents} propose an end-to-end multi-turn RL framework for web agents, achieving higher task success rates by optimizing agent decisions over long-horizon interactions.
However, the use of RL for simulating step-by-step personalized user behaviors remains underexplored. This presents a significant opportunity for future work.

\section{Methods}
\subsection{Task Formulation}

Following the setup in the OPeRA dataset~\citep{wang2025operadatasetobservationpersona}, we formulate the objective as a next-action prediction task. Given a shopping session $j$, the model observes a history of user actions $\{a_1, a_2, ..., a_{t-1}\}$, their associated rationales (i.e., Why does user conduct this specific action) $\{r_1, r_2, ..., r_{t-1}\}$, the sequence of web observations (i.e., HTML states) $\{o_1, o_2, ..., o_t\}$, and a user persona $P_i$. The model is tasked with generating the rationale $r_t$ for the next action and predicting the next action $a_t$. Formally, the learning objective is to model the function:
$$
r_t, a_t = F(a_{1...{t-1}}, r_{1...{t-1}}, o_{1...t}, P_i)
$$
Each action type is associated with specific attributes that must also be predicted. Table~\ref{tab:action_schema} summarizes the required attributes for each action type.

\begin{table*}[th]
\centering
\begin{tabular}{lll}
\toprule
\textbf{Action Type} & \textbf{Attributes} & \textbf{Example} \\
\midrule
click & element\_name & click on ``filter\_price'' \\
input & element\_name, text & type ``earbuds'' into the search box \\
terminate & None & terminate the session\\
\bottomrule
\end{tabular}
\caption{Required prediction attributes and example for each action type.}
\label{tab:action_schema}
\end{table*}

\subsection {\projectname Framework} 
The \projectname framework is illustrated in Figure~\ref{fig:framework}. Each simulation step is grounded in a real person from the dataset, with a corresponding action history, web page HTML, and annotated reasoning steps. The model is instructed to generate a rationale for conducting an immediate next action as well as the corresponding action.

We incorporate a rich user persona consists of surveys and interviews, which capture user demographics, personality traits, and shopping preferences. These persona profiles provide high-level behavioral tendencies (e.g., brand loyalty, price sensitivity) that help the model generate actions consistent with an individual's style rather than an``average'' user. Grounding in real-person personas allows the simulation to reproduce authentic decision patterns that are often missing from generic user models. 

In the prompt, we explicitly inject the persona description and instruct the model to follow it when producing plausible next actions. Nevertheless, the simulation remains context-driven: if the persona conflicts with evidence from the current page or the user’s task goals, the latter take precedence to maintain realism and task coherence.

To optimize the model, we define a verifiable reward function based on the predicted action. Specifically, we introduce a two-part reward computation:

\noindent\textbf{1) Action reward} $R_{\text{action}}$: This component measures the correctness of the predicted action by directly comparing it against the ground truth action. The reward is given only when both the action type and all required action attributes match exactly.

\begin{equation}
    R_{\text{action}} = 
    \begin{cases}
        1 & \text{if } \hat{a}_{\text{type}} = a^*_{\text{type}} \ \text{and} \ \hat{a}_{\text{attr}} = a^*_{\text{attr}} \\
        0 & \text{otherwise}
    \end{cases}
\end{equation}
where $\hat{a}$ is the predicted action, and $a^*$ is the ground-truth action (As shown in Table~\ref{tab:action_schema}).
A reward of 1 is assigned only if all required fields match exactly between prediction and ground truth. For example, for a click action, the model needs to predict both the action type as well as the clicked element name correct.

\noindent\textbf{2) Format reward} $R_{\text{format}}$: This binary reward ensures that the predicted action follows a predefined JSON schema that regulates generated reasoning and action.

The overall reward is computed as:
\begin{equation}
    R = w(\hat{a})\cdot R_{\text{action}}+R_{\text{format}} 
\end{equation}
where $\hat{a}$ is the predicted action.
Given that simulating user behavior is a challenging generation task~\citep{wang2025operadatasetobservationpersona}, we introduce a pre-defined difficulty-aware weighting function $w(a)$ that amplifies the reward for correctly predicting complex actions. This design mitigates the model’s tendency to overfit to frequent, simple actions and incentivizes accurate prediction of rarer but more informative behaviors. In addition, the output format enforces the model to first generate a rationale before the action, which implicitly guides the model toward generating a more informative reasoning process alongside correct actions.

We adopt the Group Relative Policy Optimization (GRPO)~\citep{deepseekai2025deepseekr1incentivizingreasoningcapability} method as the reinforcement learning objective. The overall optimize goal is as follows:

\begin{equation}
\begin{aligned}
&J(\theta)
= \mathbb{E}\Bigg[
 \frac{1}{G}\sum_{i=1}^{G}\frac{1}{|o_i|}\sum_{t=1}^{|o_i|}
 \min\Big(
   r_{i,t}(\theta)\,\tilde A_i,\;
   \operatorname{clip}\\&\big(r_{i,t}(\theta),\,1-\varepsilon,\,1+\varepsilon\big)\,\tilde A_i
 \Big)\quad -\beta\, D_{\mathrm{KL}}\!\big(\pi_\theta \,\Vert\, \pi_{\mathrm{ref}}\big)
\Bigg].
\end{aligned}
\end{equation}

Here, $r_{i,t}$ is the ratio between the new and old policy probabilities for sample $i$ at token $t$, and $\tilde A_i$ is the group-relative advantage:
\begin{equation}
r_{i,t}(\theta)
=\frac{\pi_\theta\!\left(o_{i,t}\mid q, o_{i,<t}\right)}
       {\pi_{\theta_{\mathrm{old}}}\!\left(o_{i,t}\mid q, o_{i,<t}\right)} .
\end{equation}

\begin{equation}
\tilde A_i
=\frac{R_i-\mu_R}{\sigma_R+\delta},\qquad
\end{equation}

\section{Experiments}
\subsection{Data Processing}
We conduct experiments on the OPeRA-filtered dataset~\citep{wang2025operadatasetobservationpersona}, the only publicly available real-world dataset for verifiable, step-wise human behavior simulation, which contains 527 real-world online shopping sessions, comprising 5,856 \textless action, observation\textgreater\ pairs and 207 annotated rationales from 49 real users.
The overall action distribution is shown in Table~\ref{tab:action_dist}. On average, each session contains 11.11 actions. Among them, \texttt{click} is the dominant action type and is further splitted into 13 fine-grained subtypes as shown in Table~\ref{tab:click_subtypes}. All sessions either end with a \texttt{click on purchase related button} action or a \texttt{terminate} action.

\begin{table}[th]
  \centering
  \begin{tblr}{
    colspec={lcc},
    row{1}={font=\bfseries}
  }
    \toprule
    Action Type & Count & Percentage \\
    \midrule
    Click     & 5{,}051 & 86.3\% \\
    Input     & 597     & 10.2\% \\
    Terminate & 208     & 3.6\%  \\
    \midrule
    All       & 5{,}856 & -- \\
    \bottomrule
  \end{tblr}
  \caption{Action in OPeRA-filtered.}
  \label{tab:action_dist}
\end{table}

Given the long HTML-based contexts in these sessions, we implement a dynamic content selection strategy to fit within the model’s maximum context length $N$. For each input, we compute its token length $L$; if $L > N$, we iteratively truncate by discarding the earliest HTMLs while preserving the full HTML content for the most recent interactions. For older interactions, we keep only the action and rationale tokens. This preserves temporally relevant page context while retaining semantically rich behavioral cues from earlier actions.

\begin{table}[t]
\centering
\begin{tblr}{
  colspec={lcc},
  row{1}={font=\bfseries},
}
\toprule
\textbf{Click Type} & \textbf{Count} & \textbf{Percentage} \\
\hline
review             & 1052 & 20.8\% \\
search             & 763  & 15.1\% \\
product\_option     & 700  & 13.9\% \\
product\_link       & 537  & 10.6\% \\
other              & 449  & 8.9\%  \\
purchase           & 321  & 6.4\%  \\
nav\_bar            & 283  & 5.6\%  \\
page\_related       & 198  & 3.9\%  \\
quantity           & 191  & 3.8\%  \\
suggested\_term     & 182  & 3.6\%  \\
cart\_side\_bar      & 145  & 2.9\%  \\
cart\_page\_select   & 139  & 2.8\%  \\
filter             & 91   & 1.8\%  \\
\bottomrule
\end{tblr}
\caption{Fine-Grained Click Type distribution in OPeRA-filtered dataset.}
\label{tab:click_subtypes}
\end{table}

In addition, the modeling framework requires the model to generate a rationale alongside each predicted action. However, as some action entries in the dataset lack annotated rationales, directly training with such incomplete data would hinder supervised fine-tuning (SFT). To address this, we adopt the rationale augmentation approach proposed by~\citet{lu2025prompting}. Specifically, we employ \texttt{claude-3.5-sonnet} to generate synthetic rationales by conditioning on the current HTML context and the user’s executed action with real-user examples. These model-generated rationales serve as plausible supervisory signals.

\begin{table*}[th]
\centering
\begin{tblr}{
  colspec={c c c c c c},
  cells={font=},
  row{1}={font=\bfseries},
}
\toprule
Method &
\makecell{Next Action Gen.\\(Accuracy)} &
\makecell{Action Type\\(Macro-F1)} &
\makecell{Fine-grained Type\\(Accuracy)} &
\makecell{Session Outcome\\(Weighted-F1)}
\\
\midrule
 Zero-shot Inference & 7.32 & 33.43&25.72&41.11\\
 RL   & 24.72   & 31.17 & 39.58& 40.51\\
SFT     &    35.14   & 72.66 & 56.43 & 66.29\\
 SFT+RL & \textbf{39.58} &\textbf{78.50} &\textbf{61.20}& \textbf{79.45}\\
\bottomrule
\end{tblr}
\caption{Evaluation of next action prediction task. `Next Action Gen.': Next Action Generation. All metrics are reported as percentages (\%). }
\label{tab:action_predict}
\end{table*}

\subsection{Evaluation}
 To evaluate model performance on the next action prediction task, we utilize the following evaluation metrics:
\textbf{a) Next Action Generation Accuracy}: An action prediction is considered correct only when all required components exactly match the ground truth. For example, for input actions, the model need to correctly predict the action type, input area, as well as the input text;
\textbf{b) Action Type F1}: Measures the correctness of action type classification. Given the highly unbalanced action type distribution, the macro F1 score is reported.
\textbf{c) Fine-grained Type Accuracy}: This metric measures the accuracy of predicted action types with finer granularity. For click actions, we first calculate the click subtype from the predicted target element and then compare it to the ground-truth. For non-click actions, we assess whether the model correctly identifies them as \texttt{terminate} or \texttt{input}. This provides a more detailed view of the model's understanding of user behavior patterns.
\textbf{d) Session Outcome F1}: Evaluates whether the session is correctly predicted to end in \texttt{click on purchase related button} or \texttt{terminate}, capturing the overall user intent.
These metrics collectively reflect both the step-wise fidelity of behavior prediction and the ultimate decision quality of the simulated user.

\subsection{Experimental Setup}
We use \texttt{Qwen2.5-7B-Instruct-1M}~\cite{yang2025qwen251mtechnicalreport} as the main model and experiment with four training configurations. In the \textbf{Zero-shot Inference} setting, the model generates actions directly without any task-specific fine-tuning. In the \textbf{SFT} setting, we apply supervised fine-tuning using behavior traces annotated with ground-truth actions. In the \textbf{RL} setting, the model is optimized via GRPO with verifiable action-level rewards. Finally, in the \textbf{SFT+RL} setting, reinforcement learning is initialized from the SFT checkpoint to improve the training stability.

In terms of reward weighting, the action reward ($R_{\text{action}}$) is scaled by task difficulty in the SFT+RL setting: 
a) correct prediction of text inputs receive 2000; 
b) correct prediction on most click types (harder click subtypes) receive 1000; 
c) correct prediction of clicks on product\_option receive 10; 
d) correct predicting clicks on reviews or search button receive 1; 
e) termination receives 1; 
f) incorrect clicks receive $-1$. 
In the RL-only setting, we use the same weighting scheme except that incorrect clicks receive $0$ instead of a negative reward, since negative rewards made training unstable.

SFT training uses a standard token-level cross-entropy objective with the AdamW optimizer, a base learning rate of $1\times10^{-5}$, 150 warm-up steps, and 2,000 total training steps with a batch size of 64. RL training is conducted using the VERL + Megatron framework, with tensor model parallelism, context parallelism, and activation checkpointing enabled. We train for 2 epochs with a batch size of 64. All experiments are conducted on $8\times8$ P4de clusters, each equipped with A100 (80 GB) GPUs.

The prompt used is shown in Appendix~\ref{prompt}.
\subsection{Main Results} 
Table~\ref{tab:action_predict} presents the results for the next-action prediction task across all four settings. Zero-shot performance is low, with an Next Action Generation Accuracy of only 7.32\%. This highlights the difficulty of behavioral prediction and the limitations of relying on pretrained knowledge alone without model adaptation. RL training alone improves exact match accuracy to 24.72\%,  but is unstable across other metrics. Applying SFT leads to significant improvements, boosting Next Action Generation Accuracy to 35.14\%, and substantially improving both Action-Type F1 score and Fine-Grained Type Accuracy (72.66\% and 56.43\%, respectively). 
Combining SFT with RL yields the best performance across all metrics. By first grounding the model with supervised learning and then applying RL-based optimization initialized from the SFT checkpoint, this setting benefits from both stable pretraining and reward-driven refinement. Specifically, the method achieves the highest Next Action Generation accuracy of 39.58\%, the best Macro F1 score, Fine-Grained Type Accuracy (78.50\% and 61.20\% respectively), and the highest Session Outcome F1 score of 79.45\%.

\subsection{Effect of Persona and Rationale}

\begin{table*}[th]
\centering
\begin{tblr}{
  colspec={l l  c c c c},
  row{1}={font=\bfseries},
}
\toprule
Method & Setting & \makecell{Next Action Gen.\\(Accuracy)} &  \makecell{Action Type\\(Macro-F1)} &\makecell{Fine-grained Type\\(Accuracy) }&\makecell{Session Outcome\\(Weighted-F1)} \\
\midrule
Zero-shot&-& 7.32 & 33.43 & 25.72 &41.11\\
Zero-shot& w/o persona & 10.20   & 33.10 &26.05 &35.88\\
Zero-shot& \hspace{0.5em}w/o rationale & 4.10   & 25.33 &16.91 &38.78\\

RL &-& 24.72 & 31.17& 39.58& 40.51\\
RL & w/o persona & 26.27 &31.20 & 41.13& 32.46\\
RL & \hspace{0.5em}w/o rationale& 12.64  & 31.20 &20.84 &44.25\\

SFT &- &35.14&75.28 & 56.43&75.85\\
SFT & w/o persona & 35.37  & 64.22 & 57.43 & 60.95 \\
SFT & \hspace{0.5em}w/o rationale &32.04 & 67.93 & 52.22&71.38\\
SFT+RL &-& \textbf{39.58} &\textbf{78.50} & \textbf{61.20} & \textbf{79.45}\\
SFT+RL& w/o persona  & 37.80 & 66.67 &59.42 &59.73\\
SFT+RL& \hspace{0.5em}w/o rationale & 34.15&73.15 &53.99&67.37\\
\bottomrule
\end{tblr}
\caption{Model performance without persona or rationale. `Next Action Gen.': Next Action Generation, `Zero-shot': Zero-shot Inference. All metrics are reported as percentages (\%). }
\label{tab:ablation_p}
\end{table*}

We quantify how explicit \emph{persona} and intermediate \emph{rationale} affect personalized behavior. Table~\ref{tab:ablation_p} ablates these signals under four training regimes (Zero-shot, RL, SFT, SFT+RL) by removing persona text from the prompt (\emph{w/o persona}) and further removing rationale from both input and generation (\emph{w/o rationale}).

Under \textbf{SFT+RL}, both signals matter and they are complementary. Removing persona reduces Next Action Generation Accuracy by 1.78 points (39.58$\rightarrow$37.80), Macro-F1 by 11.83 (78.50$\rightarrow$66.67), Fine-grained Type Accuracy by 1.78 (61.20$\rightarrow$59.42), and Session Outcome F1 by 19.72 (79.45$\rightarrow$59.73). This indicates that persona provides user-level priors that help balance action types and decide when to purchase versus terminate. Removing rationale also consistently hurts performance. This shows that step-wise reasoning supports precise behavior generation.

For \textbf{Zero-shot} and \textbf{RL-only}, removing persona can increase some surface metrics (e.g., Next Action Generation Accuracy), likely because these weaker models are not trained to use long persona text and the extra input may act as noise. However, Outcome F1 often degrades (e.g., Zero-shot 41.11$\rightarrow$35.88, RL 40.51$\rightarrow$32.46), suggesting that ignoring persona compromises user-level intent. In \textbf{SFT}, Next Action Generation Accuracy is similar with or without persona (35.14 vs.\ 35.37), but Macro-F1 and Outcome already show clear gains with persona (64.22 $\rightarrow$75.28 and 60.95$\rightarrow$75.85), implying that persona mainly helps general type balance and end-of-session decisions. Across all regimes, removing rationale is harmful, confirming that rationales act as a scaffold tying local page context to the chosen action.

\begin{table*}[th]
\centering
\begin{tblr}{
  colspec={lcccccc},
  row{1}={font=\bfseries},
}
\toprule
Model Size & Context & \makecell{Next Action Gen.\\(Accuracy)} &\makecell{Action Type\\(Macro-F1)}&\makecell{Fine-grained Type\\(Accuracy)} &\makecell{Session Outcome\\(Weighted-F1)} \\
\midrule
Qwen2.5-7B & 65k & 24.72    &31.17 &39.58 & 40.51\\
Qwen2.5-7B & 40k & 18.85& 31.14 & 28.60&41.41\\
Qwen2.5-3B & 65k & 18.07  & 31.30 &38.91&3.97\\
\bottomrule
\end{tblr}
\caption{Ablation results showing the effect of model size and context length. `Next Action Gen.': Next Action Generation. All metrics are reported as percentages (\%). }
\label{tab:ablation_m}
\end{table*}

\subsection{Effect of Model Size and Context Length}
We further experiment with a smaller backbone model, \texttt{Qwen2.5-3B-Instruct}~\cite{qwen2025qwen25technicalreport}, and compare two context length settings (40k vs. 65k tokens) under the reinforcement learning setup. As shown in Table~\ref{tab:ablation_m}, we observe clear performance improvements when increasing the context length for the 7B model. Specifically, Next Action Generation Accuracy rises from 18.85\% to 24.72\%, and Fine-Grained Type Accuracy improves from 28.60\% to 39.58\%. Although the session outcome metric slightly decreases, this is primarily due to the 40k context model overfitting on the ``click on purchase button'' action, leading to inflated outcome predictions.  The longer context window provides more examples of how a user would react to certain context and helps the model better retain earlier user intents, which is crucial for accurate simulation of behavior. 
In addition, the 3B model performs significantly worse across all metrics. Its Next Action Generation Accuracy drops to 18.07\%, and most notably, the Session Outcome F1 score falls sharply to just 3.97\%. This indicates that the smaller model fails to capture complex user intent.

\begin{table*}[ht]
\centering
\begin{tblr}{
colspec={l c c c c},
row{1} = {font=\bfseries},
}
\toprule
Action Type & Ground Truth & Predicted & Correct & Accuracy \\
\midrule
Click & 786 & 831 & 739 & 94.0\% \\
Terminate & 40 & 0 & 0 & 0.0\% \\
Input & 76 & 4 & 1 & 1.3\% \\
Other & 0 & 67 & 0 & 0.0\% \\
\bottomrule
\end{tblr}
\caption{RL-only action type distribution and accuracy (i.e. Reward Hacking). The model ignores \texttt{input}/\texttt{terminate} and produces spurious \texttt{other} actions.}
\label{tab:type_rl}
\end{table*}

\begin{table*}[th]
\centering
\begin{tblr}{
  colspec={l l  c c c c},
  row{1}={font=\bfseries},
}
\toprule
Method & Setting & \makecell{Next Action Gen.\\(Accuracy)} &  \makecell{Action Type\\(Macro-F1)} &\makecell{Fine-grained Type\\(Accuracy) }&\makecell{Session Outcome\\(Weighted-F1)} \\
\midrule

SFT+RL &persona& \textbf{39.58} &\textbf{78.50} & \textbf{61.20} & \textbf{79.45}\\
SFT+RL& shuffle & 28.94 & 38.88 & 40.35 & 48.41\\

\bottomrule
\end{tblr}
\caption{Model performance after shuffle the persona.'Zero-shot': Zero-shot Inference. All metrics are reported as percentages (\%). }
\label{tab:ablation_shuffle}
\end{table*}

\subsection{Analysis}

We further analyze the model's error patterns under different training regimes to understand its behavior and limitations.

\paragraph{From Reward Hacking to Balance with SFT Init.}
Without supervised grounding, the RL-only model learns to exploit the reward by favoring frequent, simple moves. As Table~\ref{tab:type_rl} shows, the model mostly predicts \texttt{click} action type and never predicts \texttt{input} or \texttt{terminate}.

Moreover, we noticed that under RL-only setting, the model over-selects subtypes with strong surface cues (e.g., \texttt{purchase}, \texttt{review}, \texttt{search}). While these predictions result in superficially high reward, they fail to reflect the diversity of real user behavior. This highlights a core limitation of naive reward-driven optimization: the learned policy may appear effective but lacks true generalization capability. 

In contrast, initializing RL from a SFT checkpoint breaks this shortcut. The SFT policy already assigns non-trivial probability to rare actions, so RL can refine a balanced policy instead of relearning from scratch. The model shows a more balanced action prediction distribution and recovery of underrepresented actions (e.g., \texttt{terminate}) 
(Detail action distribution can be found in Appendix~\ref{sec:appendix}).

\paragraph{Persona Guides How to Act and When to Stop.}
Within SFT+RL, real-person personas provide user-level priors (e.g., price sensitivity, brand loyalty) that resolve ties when page evidence alone is ambiguous. Removing persona destabilizes action type prediction and weakens end-of-session decisions. Specifically, the model shows a drift with fewer correct \texttt{purchase} and \texttt{terminate} predictions (Appendix~\ref{sec:appendix}). 
In addition, Table~\ref{tab:ablation_shuffle} shows the results after shuffling the persona in the input information. Breaking the alignment between the user and their profile causes large drops across all metrics: Next Action Generation Accuracy 39.58$\rightarrow$28.94, Action Type Macro-F1 78.50$\rightarrow$38.88, Fine-grained Type Accuracy 61.20$\rightarrow$40.35, and Session Outcome Weighted-F1 79.45$\rightarrow$48.41.  Moreover, persona information increases recall of rare but consequential actions (especially \texttt{terminate}) and improves calibration across types, yielding higher overall performance. 

All these results demonstrates that persona information shifts the model policy from an \emph{average user} heuristic to \emph{this specific user’s} behavior, deciding which element \emph{this user} would act on and whether \emph{this user} would stop. 

\section{Conclusion}
We introduced \projectname, a reinforcement learning method for step-wise, personalized user behavior simulation in online shopping. By conditioning the policy on explicit persona information and optimizing a tailored reward that favors action correctness, the model achieves superior next-action prediction accuracy on the OPeRA dataset and stronger personalization than prompting and SFT baselines. Comprehensive ablations and analysis show that persona and rationale make complementary contributions: persona supplies user-level priors that guide action selection under ambiguous page states, while rationale supports more stable credit assignment during RL. This combination improves calibration across action types and increases recall of rare but consequential actions such as \texttt{terminate}, leading to better end-of-session decisions and overall performance.

\section*{Limitations}
Despite the performance gains, we observe that the policy still shows a bias toward frequent, simple actions and can under-predict infrequent, user-specific intents. Our reward focuses on correctness at the action level and does not fully capture user satisfaction or effort. Future work could explore richer, more localized reward signals and context, together with stronger persona representations and integration methods.

\bibliography{custom}

\begin{thebibliography}{30}
\providecommand{\natexlab}[1]{#1}

\bibitem[{Aher et~al.(2023)Aher, Arriaga, and Kalai}]{10.5555/3618408.3618425}
Gati Aher, Rosa~I. Arriaga, and Adam~Tauman Kalai. 2023.
\newblock Using large language models to simulate multiple humans and replicate
  human subject studies.
\newblock In \emph{Proceedings of the 40th International Conference on Machine
  Learning}, ICML'23. JMLR.org.

\bibitem[{Anthropic(2025)}]{anthropic2025claude37}
Anthropic. 2025.
\newblock \href {https://www.anthropic.com/news/claude-3-7-sonnet} {Claude 3.7
  sonnet and claude code}.
\newblock Accessed: 2025-04-06.

\bibitem[{Chen et~al.(2025)Chen, Tu, Wang, Liu, Tang, Du, Zhou, and
  Xie}]{chen2025sft}
Hardy Chen, Haoqin Tu, Fali Wang, Hui Liu, Xianfeng Tang, Xinya Du, Yuyin Zhou,
  and Cihang Xie. 2025.
\newblock Sft or rl? an early investigation into training r1-like reasoning
  large vision-language models.
\newblock \emph{arXiv preprint arXiv:2504.11468}.

\bibitem[{DeepSeek-AI et~al.(2025)DeepSeek-AI, Guo, Yang, Zhang, Song, Zhang,
  Xu, Zhu, Ma, Wang, Bi, Zhang, Yu, Wu, Wu, Gou, Shao, Li, Gao, Liu, Xue, Wang,
  Wu, Feng, Lu, Zhao, Deng, Zhang, Ruan, Dai, Chen, Ji, Li, Lin, Dai, Luo, Hao,
  Chen, Li, Zhang, Bao, Xu, Wang, Ding, Xin, Gao, Qu, Li, Guo, Li, Wang, Chen,
  Yuan, Qiu, Li, Cai, Ni, Liang, Chen, Dong, Hu, Gao, Guan, Huang, Yu, Wang,
  Zhang, Zhao, Wang, Zhang, Xu, Xia, Zhang, Zhang, Tang, Li, Wang, Li, Tian,
  Huang, Zhang, Wang, Chen, Du, Ge, Zhang, Pan, Wang, Chen, Jin, Chen, Lu,
  Zhou, Chen, Ye, Wang, Yu, Zhou, Pan, Li, Zhou, Wu, Ye, Yun, Pei, Sun, Wang,
  Zeng, Zhao, Liu, Liang, Gao, Yu, Zhang, Xiao, An, Liu, Wang, Chen, Nie,
  Cheng, Liu, Xie, Liu, Yang, Li, Su, Lin, Li, Jin, Shen, Chen, Sun, Wang,
  Song, Zhou, Wang, Shan, Li, Wang, Wei, Zhang, Xu, Li, Zhao, Sun, Wang, Yu,
  Zhang, Shi, Xiong, He, Piao, Wang, Tan, Ma, Liu, Guo, Ou, Wang, Gong, Zou,
  He, Xiong, Luo, You, Liu, Zhou, Zhu, Xu, Huang, Li, Zheng, Zhu, Ma, Tang,
  Zha, Yan, Ren, Ren, Sha, Fu, Xu, Xie, Zhang, Hao, Ma, Yan, Wu, Gu, Zhu, Liu,
  Li, Xie, Song, Pan, Huang, Xu, Zhang, and
  Zhang}]{deepseekai2025deepseekr1incentivizingreasoningcapability}
DeepSeek-AI, Daya Guo, Dejian Yang, Haowei Zhang, Junxiao Song, Ruoyu Zhang,
  Runxin Xu, Qihao Zhu, Shirong Ma, Peiyi Wang, Xiao Bi, Xiaokang Zhang,
  Xingkai Yu, Yu~Wu, Z.~F. Wu, Zhibin Gou, Zhihong Shao, Zhuoshu Li, Ziyi Gao,
  and 181 others. 2025.
\newblock \href {https://arxiv.org/abs/2501.12948} {Deepseek-r1: Incentivizing
  reasoning capability in llms via reinforcement learning}.
\newblock \emph{Preprint}, arXiv:2501.12948.

\bibitem[{Helmi et~al.(2023)Helmi, Komaladewi, Sarasi, and
  Yolanda}]{helmi2023characterizing}
Arief Helmi, Rita Komaladewi, Vita Sarasi, and Ledy Yolanda. 2023.
\newblock Characterizing young consumer online shopping style: Indonesian
  evidence.
\newblock \emph{Sustainability}, 15(5):3988.

\bibitem[{Jin et~al.(2025)Jin, Zeng, Yue, Yoon, Arik, Wang, Zamani, and
  Han}]{jin2025search}
Bowen Jin, Hansi Zeng, Zhenrui Yue, Jinsung Yoon, Sercan Arik, Dong Wang, Hamed
  Zamani, and Jiawei Han. 2025.
\newblock Search-r1: Training llms to reason and leverage search engines with
  reinforcement learning.
\newblock \emph{arXiv preprint arXiv:2503.09516}.

\bibitem[{Lin et~al.(2025)Lin, Wang, and Qian}]{lin2025rec}
Jiacheng Lin, Tian Wang, and Kun Qian. 2025.
\newblock Rec-r1: Bridging generative large language models and user-centric
  recommendation systems via reinforcement learning.
\newblock \emph{arXiv preprint arXiv:2503.24289}.

\bibitem[{Lu et~al.(2025{\natexlab{a}})Lu, Huang, Han, Yao, Bei, Gesi, Xie, He,
  Wang et~al.}]{lu2025prompting}
Yuxuan Lu, Jing Huang, Yan Han, Bingsheng Yao, Sisong Bei, Jiri Gesi, Yaochen
  Xie, Qi~He, Dakuo Wang, and 1 others. 2025{\natexlab{a}}.
\newblock Prompting is not all you need! evaluating llm agent simulation
  methodologies with real-world online customer behavior data.
\newblock \emph{arXiv preprint arXiv:2503.20749}.

\bibitem[{Lu et~al.(2025{\natexlab{b}})Lu, Yao, Gu, Huang, Wang, Li, Gesi, He,
  Li, and Wang}]{lu2025uxagentllmagentbasedusability}
Yuxuan Lu, Bingsheng Yao, Hansu Gu, Jing Huang, Jessie Wang, Laurence Li, Jiri
  Gesi, Qi~He, Toby Jia-Jun Li, and Dakuo Wang. 2025{\natexlab{b}}.
\newblock \href {https://arxiv.org/abs/2502.12561} {Uxagent: An llm agent-based
  usability testing framework for web design}.
\newblock \emph{Preprint}, arXiv:2502.12561.

\bibitem[{McConnell(1974)}]{mcconnell1974understanding}
James~V McConnell. 1974.
\newblock \emph{Understanding human behavior: An introduction to psychology.}
\newblock Holt, Rinehart \& Winston.

\bibitem[{Ouyang et~al.(2022)Ouyang, Wu, Jiang, Almeida, Wainwright, Mishkin,
  Zhang, Agarwal, Slama, Ray, Schulman, Hilton, Kelton, Miller, Simens, Askell,
  Welinder, Christiano, Leike, and
  Lowe}]{ouyang2022traininglanguagemodelsfollow}
Long Ouyang, Jeff Wu, Xu~Jiang, Diogo Almeida, Carroll~L. Wainwright, Pamela
  Mishkin, Chong Zhang, Sandhini Agarwal, Katarina Slama, Alex Ray, John
  Schulman, Jacob Hilton, Fraser Kelton, Luke Miller, Maddie Simens, Amanda
  Askell, Peter Welinder, Paul Christiano, Jan Leike, and Ryan Lowe. 2022.
\newblock \href {https://arxiv.org/abs/2203.02155} {Training language models to
  follow instructions with human feedback}.
\newblock \emph{Preprint}, arXiv:2203.02155.

\bibitem[{Park et~al.(2023)Park, O'Brien, Cai, Morris, Liang, and
  Bernstein}]{park2023generative}
Joon~Sung Park, Joseph O'Brien, Carrie~Jun Cai, Meredith~Ringel Morris, Percy
  Liang, and Michael~S Bernstein. 2023.
\newblock Generative agents: Interactive simulacra of human behavior.
\newblock In \emph{Proceedings of the 36th annual acm symposium on user
  interface software and technology}, pages 1--22.

\bibitem[{Park et~al.(2024)Park, Zou, Shaw, Hill, Cai, Morris, Willer, Liang,
  and Bernstein}]{park2024generative}
Joon~Sung Park, Carolyn~Q Zou, Aaron Shaw, Benjamin~Mako Hill, Carrie Cai,
  Meredith~Ringel Morris, Robb Willer, Percy Liang, and Michael~S Bernstein.
  2024.
\newblock Generative agent simulations of 1,000 people.
\newblock \emph{arXiv preprint arXiv:2411.10109}.

\bibitem[{Qwen et~al.(2025)Qwen, :, Yang, Yang, Zhang, Hui, Zheng, Yu, Li, Liu,
  Huang, Wei, Lin, Yang, Tu, Zhang, Yang, Yang, Zhou, Lin, Dang, Lu, Bao, Yang,
  Yu, Li, Xue, Zhang, Zhu, Men, Lin, Li, Tang, Xia, Ren, Ren, Fan, Su, Zhang,
  Wan, Liu, Cui, Zhang, and Qiu}]{qwen2025qwen25technicalreport}
Qwen, :, An~Yang, Baosong Yang, Beichen Zhang, Binyuan Hui, Bo~Zheng, Bowen Yu,
  Chengyuan Li, Dayiheng Liu, Fei Huang, Haoran Wei, Huan Lin, Jian Yang,
  Jianhong Tu, Jianwei Zhang, Jianxin Yang, Jiaxi Yang, Jingren Zhou, and 25
  others. 2025.
\newblock \href {https://arxiv.org/abs/2412.15115} {Qwen2.5 technical report}.
\newblock \emph{Preprint}, arXiv:2412.15115.

\bibitem[{Rafailov et~al.(2024)Rafailov, Sharma, Mitchell, Ermon, Manning, and
  Finn}]{rafailov2024directpreferenceoptimizationlanguage}
Rafael Rafailov, Archit Sharma, Eric Mitchell, Stefano Ermon, Christopher~D.
  Manning, and Chelsea Finn. 2024.
\newblock \href {https://arxiv.org/abs/2305.18290} {Direct preference
  optimization: Your language model is secretly a reward model}.
\newblock \emph{Preprint}, arXiv:2305.18290.

\bibitem[{Schulman et~al.(2017)Schulman, Wolski, Dhariwal, Radford, and
  Klimov}]{schulman2017proximalpolicyoptimizationalgorithms}
John Schulman, Filip Wolski, Prafulla Dhariwal, Alec Radford, and Oleg Klimov.
  2017.
\newblock \href {https://arxiv.org/abs/1707.06347} {Proximal policy
  optimization algorithms}.
\newblock \emph{Preprint}, arXiv:1707.06347.

\bibitem[{Shao et~al.(2024)Shao, Wang, Zhu, Xu, Song, Bi, Zhang, Zhang, Li, Wu
  et~al.}]{shao2024deepseekmath}
Zhihong Shao, Peiyi Wang, Qihao Zhu, Runxin Xu, Junxiao Song, Xiao Bi, Haowei
  Zhang, Mingchuan Zhang, YK~Li, Yang Wu, and 1 others. 2024.
\newblock Deepseekmath: Pushing the limits of mathematical reasoning in open
  language models.
\newblock \emph{arXiv preprint arXiv:2402.03300}.

\bibitem[{Shinn et~al.(2023)Shinn, Cassano, Gopinath, Narasimhan, and
  Yao}]{shinn2023reflexion}
Noah Shinn, Federico Cassano, Ashwin Gopinath, Karthik Narasimhan, and Shunyu
  Yao. 2023.
\newblock Reflexion: Language agents with verbal reinforcement learning.
\newblock \emph{Advances in Neural Information Processing Systems},
  36:8634--8652.

\bibitem[{Sreedhar et~al.(2025)Sreedhar, Cai, Ma, Nickerson, and
  Chilton}]{10.1145/3708359.3712149}
Karthik Sreedhar, Alice Cai, Jenny Ma, Jeffrey~V Nickerson, and Lydia~B
  Chilton. 2025.
\newblock \href {https://doi.org/10.1145/3708359.3712149} {Simulating
  cooperative prosocial behavior with multi-agent llms: Evidence and mechanisms
  for ai agents to inform policy decisions}.
\newblock In \emph{Proceedings of the 30th International Conference on
  Intelligent User Interfaces}, IUI '25, page 1272–1286, New York, NY, USA.
  Association for Computing Machinery.

\bibitem[{Sun et~al.(2024)Sun, Yang, Reddy, Fung, Chan, Zhai, and
  Ji}]{sun2024persona}
Chenkai Sun, Ke~Yang, Revanth~Gangi Reddy, Yi~R Fung, Hou~Pong Chan, ChengXiang
  Zhai, and Heng Ji. 2024.
\newblock Persona-db: Efficient large language model personalization for
  response prediction with collaborative data refinement.
\newblock \emph{arXiv preprint arXiv:2402.11060}.

\bibitem[{Wang et~al.(2025{\natexlab{a}})Wang, Hsu, Lu, Gu, Cui, Xie, Headean,
  Yao, Veeragouni, Liu, Nag, and Wang}]{wang2025agentabautomatedscalableweb}
Dakuo Wang, Ting-Yao Hsu, Yuxuan Lu, Hansu Gu, Limeng Cui, Yaochen Xie, William
  Headean, Bingsheng Yao, Akash Veeragouni, Jiapeng Liu, Sreyashi Nag, and
  Jessie Wang. 2025{\natexlab{a}}.
\newblock \href {https://arxiv.org/abs/2504.09723} {Agenta/b: Automated and
  scalable web a/btesting with interactive llm agents}.
\newblock \emph{Preprint}, arXiv:2504.09723.

\bibitem[{Wang et~al.(2024)Wang, Zhang, Yang, Chen, Tang, Zhang, Chen, Lin,
  Song, Zhao, Xu, Dou, Wang, and Wen}]{wang2024userbehaviorsimulationlarge}
Lei Wang, Jingsen Zhang, Hao Yang, Zhiyuan Chen, Jiakai Tang, Zeyu Zhang,
  Xu~Chen, Yankai Lin, Ruihua Song, Wayne~Xin Zhao, Jun Xu, Zhicheng Dou, Jun
  Wang, and Ji-Rong Wen. 2024.
\newblock \href {https://arxiv.org/abs/2306.02552} {User behavior simulation
  with large language model based agents}.
\newblock \emph{Preprint}, arXiv:2306.02552.

\bibitem[{Wang et~al.(2025{\natexlab{b}})Wang, Lu, Li, Amini, Sun, Bart, Lyu,
  Gesi, Wang, Huang, Su, Ehsan, Alikhani, Li, Chilton, and
  Wang}]{wang2025operadatasetobservationpersona}
Ziyi Wang, Yuxuan Lu, Wenbo Li, Amirali Amini, Bo~Sun, Yakov Bart, Weimin Lyu,
  Jiri Gesi, Tian Wang, Jing Huang, Yu~Su, Upol Ehsan, Malihe Alikhani, Toby
  Jia-Jun Li, Lydia Chilton, and Dakuo Wang. 2025{\natexlab{b}}.
\newblock \href {https://arxiv.org/abs/2506.05606} {Opera: A dataset of
  observation, persona, rationale, and action for evaluating llms on human
  online shopping behavior simulation}.
\newblock \emph{Preprint}, arXiv:2506.05606.

\bibitem[{Wei et~al.(2025)Wei, Yao, Liu, Zhang, Lu, Qiu, Yu, Xu, Zhang, Yin,
  Yun, and Li}]{wei2025webagentr1trainingwebagents}
Zhepei Wei, Wenlin Yao, Yao Liu, Weizhi Zhang, Qin Lu, Liang Qiu, Changlong Yu,
  Puyang Xu, Chao Zhang, Bing Yin, Hyokun Yun, and Lihong Li. 2025.
\newblock \href {https://arxiv.org/abs/2505.16421} {Webagent-r1: Training web
  agents via end-to-end multi-turn reinforcement learning}.
\newblock \emph{Preprint}, arXiv:2505.16421.

\bibitem[{Yang et~al.(2025)Yang, Yu, Li, Liu, Huang, Huang, Jiang, Tu, Zhang,
  Zhou, Lin, Dang, Yang, Yu, Li, Sun, Zhu, Men, He, Xu, Yin, Yu, Qiu, Ren,
  Yang, Li, Xu, and Zhang}]{yang2025qwen251mtechnicalreport}
An~Yang, Bowen Yu, Chengyuan Li, Dayiheng Liu, Fei Huang, Haoyan Huang,
  Jiandong Jiang, Jianhong Tu, Jianwei Zhang, Jingren Zhou, Junyang Lin, Kai
  Dang, Kexin Yang, Le~Yu, Mei Li, Minmin Sun, Qin Zhu, Rui Men, Tao He, and 9
  others. 2025.
\newblock \href {https://arxiv.org/abs/2501.15383} {Qwen2.5-1m technical
  report}.
\newblock \emph{Preprint}, arXiv:2501.15383.

\bibitem[{Yao et~al.(2023)Yao, Zhao, Yu, Du, Shafran, Narasimhan, and
  Cao}]{yao2023react}
Shunyu Yao, Jeffrey Zhao, Dian Yu, Nan Du, Izhak Shafran, Karthik Narasimhan,
  and Yuan Cao. 2023.
\newblock React: Synergizing reasoning and acting in language models.
\newblock In \emph{International Conference on Learning Representations
  (ICLR)}.

\bibitem[{Yu et~al.(2025)Yu, Zhang, Zhu, Yuan, Zuo, Yue, Dai, Fan, Liu, Liu,
  Liu, Lin, Lin, Ma, Sheng, Tong, Zhang, Zhang, Zhang, Zhu, Zhu, Chen, Chen,
  Wang, Yu, Song, Wei, Zhou, Liu, Ma, Zhang, Yan, Qiao, Wu, and
  Wang}]{yu2025dapoopensourcellmreinforcement}
Qiying Yu, Zheng Zhang, Ruofei Zhu, Yufeng Yuan, Xiaochen Zuo, Yu~Yue, Weinan
  Dai, Tiantian Fan, Gaohong Liu, Lingjun Liu, Xin Liu, Haibin Lin, Zhiqi Lin,
  Bole Ma, Guangming Sheng, Yuxuan Tong, Chi Zhang, Mofan Zhang, Wang Zhang,
  and 16 others. 2025.
\newblock \href {https://arxiv.org/abs/2503.14476} {Dapo: An open-source llm
  reinforcement learning system at scale}.
\newblock \emph{Preprint}, arXiv:2503.14476.

\bibitem[{Zhang et~al.(2025{\natexlab{a}})Zhang, Huang, Yao, Liu, Zhang, Lu,
  and Tao}]{zhang2025r1}
Jingyi Zhang, Jiaxing Huang, Huanjin Yao, Shunyu Liu, Xikun Zhang, Shijian Lu,
  and Dacheng Tao. 2025{\natexlab{a}}.
\newblock R1-vl: Learning to reason with multimodal large language models via
  step-wise group relative policy optimization.
\newblock \emph{arXiv preprint arXiv:2503.12937}.

\bibitem[{Zhang et~al.(2025{\natexlab{b}})Zhang, Wang, Gesi, Wang, Lu, Lin,
  Zhan, Gao, Jiao, Liu, Qian, Tang, Xue, Zhang, Cui, Guo, and
  Wang}]{zhang2025shopr1rewardingllmssimulate}
Yimeng Zhang, Tian Wang, Jiri Gesi, Ziyi Wang, Yuxuan Lu, Jiacheng Lin, Sinong
  Zhan, Vianne Gao, Ruochen Jiao, Junze Liu, Kun Qian, Yuxin Tang, Ran Xue,
  Houyu Zhang, Qingjun Cui, Yufan Guo, and Dakuo Wang. 2025{\natexlab{b}}.
\newblock \href {https://arxiv.org/abs/2507.17842} {Shop-r1: Rewarding llms to
  simulate human behavior in online shopping via reinforcement learning}.
\newblock \emph{Preprint}, arXiv:2507.17842.

\bibitem[{Zheng et~al.(2025)Zheng, Liu, Li, Chen, Yu, Gao, Dang, Liu, Men,
  Yang, Zhou, and Lin}]{zheng2025groupsequencepolicyoptimization}
Chujie Zheng, Shixuan Liu, Mingze Li, Xiong-Hui Chen, Bowen Yu, Chang Gao, Kai
  Dang, Yuqiong Liu, Rui Men, An~Yang, Jingren Zhou, and Junyang Lin. 2025.
\newblock \href {https://arxiv.org/abs/2507.18071} {Group sequence policy
  optimization}.
\newblock \emph{Preprint}, arXiv:2507.18071.

\end{thebibliography}
\clearpage
\appendix

\section{Action Distribution}
\begin{figure*}[th]
    \centering
    \includegraphics[width=0.8\linewidth]{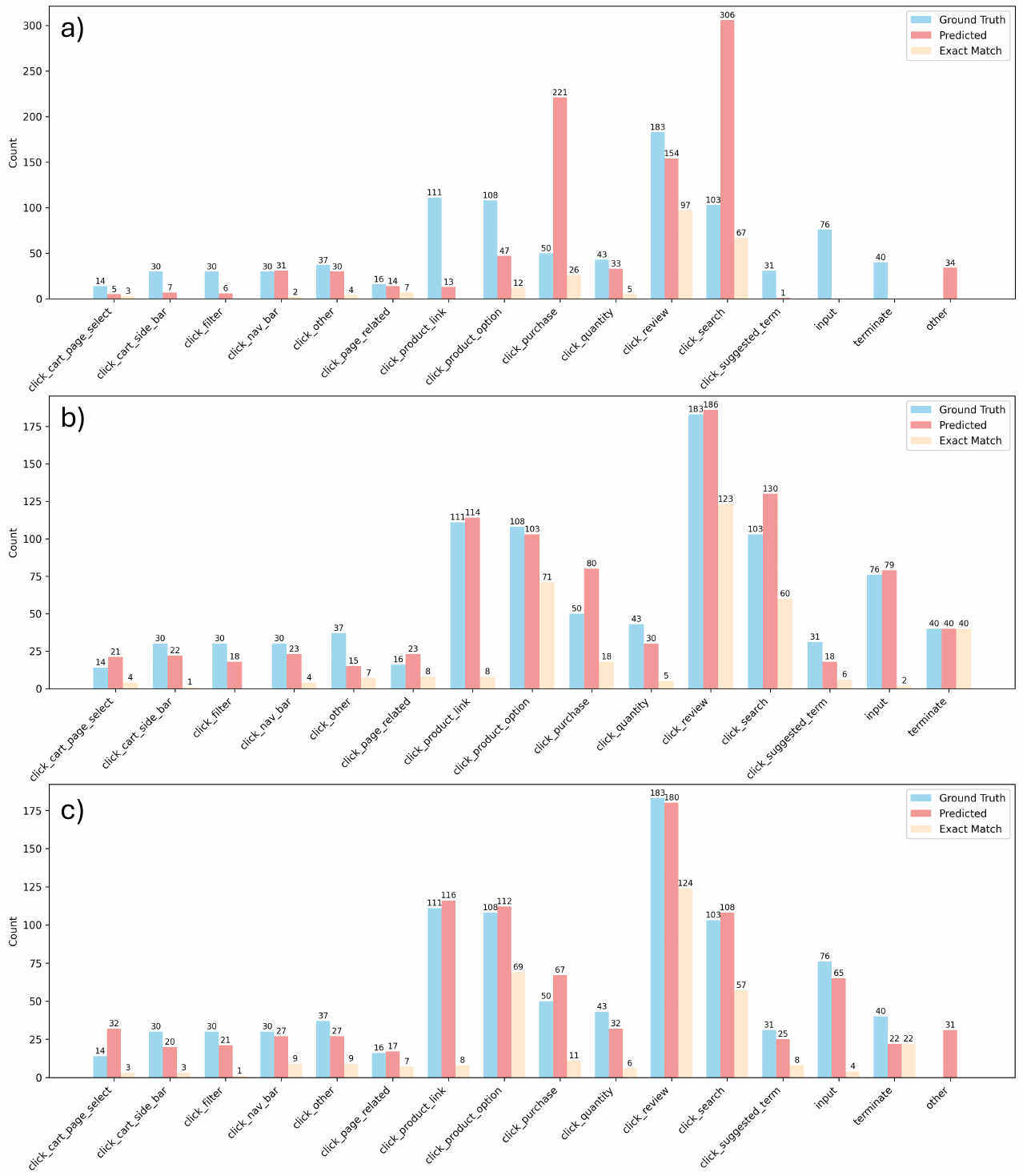}
    \caption{Fine-grained action distribution. a) Model trained using RL only. b) Model trained using SFT+RL. c) Model trained using SFT+RL without persona}
    \label{fig:fg_distribution}
\end{figure*}

Figure~\ref{fig:fg_distribution} shows the fine-grained action type distribution of ground truth actions, predicted actions, and correctly predicted actions (i.e. Exact Match) among three training regimes: a) RL-only, b) SFT+RL, c) SFT+RL without persona information. We observe that, with RL-only method, the model tends to predict mostly purchase and review and search action. In contrast, under SFT+RL setting, the policy shows a balanced predicted action distribution, while removing the persona would make the performance on predicting termination to drop.
\label{sec:appendix}

\section{Experiment Prompt Design}
\label{prompt}
Below are the two prompts for action prediction task and joint rationale and action generation task:
\begin{lstlisting}[style=promptstyle]
<IMPORTANT>
Your task is to predict the immediate next action of a shopper.
You need to pretend that you are a real user shopping on amazon.com.
The history action, rationale, context and the user persona will be provided to you.
Ensure your prediction follows natural behavior sequences (e.g., users may click a search box before typing, type a query before clicking search)
</IMPORTANT>

# Action Space

An action is represented in JSON format, and there are four primary types of actions:

#### 1. `input`:
Type text into an input field. The input field is identified by `name`.
{
    "type": "input",
    "name": "input_name",
    "text": "input_text"
}

#### 2. `click`:
Click on a button or clickable element identified by `name`.
{
    "type": "click",
    "name": "clickable_name",
}

#### 3. `terminate`:
When you are unsatisfied with the current search result and you don't want to buy anything, use `terminate` to indicate that you want to close the browser window and terminate the task.
{
    "type": "terminate"
}

# Rationale
Rationale is the reason why the user takes the action. Some of the rationale is provided to you.

# Context
Your context will be the HTML of the amazon page you are looking at. Some interactable elements will be added a unique "name" attribute, which you can use to identify the element to interact with (click or input).

# Persona
The user persona reflects the user's demographics, personality, and shopping preference. First identify which aspects of the persona might be relevant to the current shopping context, then consider them only if they naturally align with the ongoing shopping journey. DO NOT RELY ON IT.

# Output Format
You need to predict the rationale AND the corresponding next action. Your output should follow a strict JSON format:

{
    "rationale": "<rationale>", // rationale goes here, a string
    "action": {
        "type": "<type>",
        ...
    }// action goes here, a dictionary
}

<IMPORTANT>
OUTPUT A SINGLE JSON OBJECT, NOTHING ELSE.
</IMPORTANT>
\end{lstlisting}
\end{document}